%% file: ru_translate_paper.tex
\documentclass{llncs}
\usepackage{epsf}
\usepackage{url}
\usepackage{multirow}
\usepackage{float}
\usepackage{arydshln} 

\usepackage[T2A]{fontenc}
\usepackage[utf8]{inputenc}


\begin{document}

\title{Russian Language Datasets in the Digitial Humanities Domain and Their Evaluation with Word Embeddings}

\author{Gerhard Wohlgenannt$^{\ast}$, Artemii Babushkin$^{\ast}$ and Denis Romashov$^{\ast}$ and Igor Ukrainets$^{\ast}$ and Anton Maskaykin$^{\ast}$ and Ilya Shutov$^{\ast}$}

\institute{$^{\ast}$Faculty of Software Engineering and Computer Systems \\ ITMO University, St. Petersburg, Russia } 


\maketitle




%

\begin{abstract}

In this paper, we present Russian language datasets in the digital humanities domain 
for the evaluation of word embedding techniques or similar language modeling and feature learning algorithms. 
The datasets are split into two task types, word intrusion and word analogy, and contain 31362 task units in total.
The characteristics of the tasks and datasets are that they build upon small, domain-specific corpora, and that the datasets contain a high number of named entities. 
The datasets were created manually for two fantasy novel book series (``A Song of Ice and Fire'' and ``Harry Potter'').
We provide baseline evaluations with popular word embedding models trained on the book corpora for the given tasks, 
both for the Russian and English language versions of the datasets. 
Finally, we compare and analyze the results and discuss specifics of Russian language with regards to the problem setting.
 

\end{abstract}

\setlength{\textfloatsep}{15pt}
\setlength{\intextsep}{15pt}
\section{Introduction}
\label{sec:intro}

Distributional semantics base on the idea, that the meaning of a word can be estimated
from its linguistic context~\cite{Harris1954}. Recently, with the work on word2vec~\cite{mikolov2013w2v},
where prediction-based neural embedding models are trained on large corpora, word embedding models became
very popular as input to solve many natural language processing (NLP) tasks.
In word embedding models, terms are represented by low-dimensional, dense vectors
of floating-point numbers. While distributional language models are well studied in the general domain
when trained on large corpora, the situation is different regarding specialized domains, and 
term types such as \emph{proper nouns}, which exhibit specific characteristics~\cite{Herbelot2015darcy,Boleda2017eacl}.
Datasets in the digital humanities domain, which include some of these aspects, were presented by Wohlgenannt~\cite{wohlgenannt2018cicling}.
In this work, those datasets are translated to Russian language, and we provide  baseline evaluations with popular word embedding models,
and analyze differences between English and Russian experimental results. 
The manually created datasets contain \emph{analogies} and \emph{word intrusion} tasks for two popular fantasy novel book series:
``A Song of Ice and Fire'' (ASOIF, by GRR Martin) and ``Harry Potter'' (HP, by JK Rowling). The \emph{analogy} task is a well-known method for
the intrinsic evaluation of embedding models, the \emph{word intruder} task is related to word similarity and used to solve the ``odd one out''
task~\cite{gensim}.

The basic question is how well Russian language word embedding models are suited for solving such tasks, and what are the differences
to English language datasets and corpora. More specifically, what is the performance on the two task types, which word embedding algorithms
are more suitable for the tasks, and which factors are responsible for any differences between English and Russian language results?

In this work, we manually translated the datasets into Russian. In total, we present 8 datasets, for both book series, the two
task types, and the distinction between unigrams and n-gram datasets. Word2vec~\cite{mikolov2013w2v} and FastText~\cite{bojanowski2016} with different
settings were trained on the Russian (and English) book corpora, and then evaluated with the given datasets. The evaluation scores are sufficiently lower 
for Russian, but the application of lemmatization on the Russian corpora helps to partly close the gap. Other issues such as ambiguities 
in the translation of the datasets, and inconsistencies in the transliteration of English named entities are analyzed and discussed.
As an example, the best accuracy scores for the ASOIF unigram dataset for Russian are 32.7\% for the \emph{analogy} task, and
73.3\% for word intrusion, while for English the best results are 37.1\%, and 86.5\%, resp.

The main contributions include the eight Russian language datasets with 31362 task units in total, translated by two independent teams, 
baseline evaluations with various word2vec and FastText models, comparisons between English and Russian, and the analyses of the results,
specifically with regards to corpus word frequency and typical issues in translation and transliteration.

The paper is structured as follows: After an overview of related work in Section~\ref{sec:related}, the two task types and the translated datasets
are introduced in Section~\ref{sec:tasks}. Section~\ref{sec:evaluation} first elaborates the evaluation setup, for example the details of the corpora,
and the model settings used in the evaluations. Subsequently, evaluation results, both aggregated and fire-grained, are presented. We discuss the
findings in Section~\ref{sec:dissc}, and then provide conclusions in Section~\ref{sec:concl}.

%
%
%

\section{Related Work}
\label{sec:related}

Word embedding vectors are used in a many modern NLP applications to represent words,  often by 
applying pre-trained models trained on large general-purpose text corpora. Ghannay et al.~\cite{ghanny2016}
compare the performance of model types such as word2vec CBOW and skip-gram~\cite{mikolov2013w2v}, and GloVe~\cite{pennington2014glove}.
For example, FastText~\cite{bojanowski2016} provides pre-trained models for many language for download.
But there are also language-specific efforts, eg.~RusVectōrēs~\cite{kutuzov2017} include a number of models trained on various Russian corpora. Workshops
on Russian language semantic similarity~\cite{Panchenko2015RUSSETF} emphasized the importance of the research topic.
     
For the intrinsic evaluation of word embedding models, researchers often use existing word similarity datasets like
WordSim-353~\cite{finkelstein2001placing} or MEN~\cite{Bruni2014}, or analogy datasets like Google~\cite{mikolov2013w2v} or BATS~\cite{Gladkova2016bats}.

In specialized domains, large text corpora for training are often not available.
Sahlgren and Lenci~\cite{Sahlgren2016emnlp} evaluate the impact of corpus size and term frequency on accuracy in word similarity tasks, and as expected,
corpus size has a strong impact. The datasets that we translated contain a high percentage of named entities such as book 
characters and locations. Herbelot~\cite{Herbelot2015darcy} discusses various aspects of instances (like named entities) versus kinds, such
as the detection of instantiation relations in distributional models. Distributional models are shown to be better suited for categorizing than for distinguishing
individuals and their properties~\cite{Boleda2017iwcs}. This is also reflected by our results, esp.~the analysis of task difficulty
in word intrusion (see Section~\ref{sec:evalres}).

The work on using distributional methods in the digital humanities domain is limited. More efforts have been directed at 
dialog structure and social network extraction~\cite{elson2010,jayannavar2015validating} or character detection~\cite{vala2015}.



\section{Tasks and Datasets}
\label{sec:tasks}

This section introduces the task types (analogy and word intrusion), discusses the dataset translation process, and briefly describes the
word embedding algorithms used, as well as the basics of implementation.

    \subsection{Task Types}
    The datasets and evaluations focus on two task types: word intrusion and word analogy. Term analogy is a popular method
    for the evaluation of models of distributional semantics, applied for example in the original word2vec paper~\cite{mikolov2013w2v}.
    Word intrusion is a task similar to \emph{word similarity}, which is a popular intrinsic evaluation method 
    for word embedding models (see Section~\ref{sec:related}).
  
    The word analogy task captures semantic or syntactic relations between words, a well-known example is ``\emph{man} is to \emph{king}, like
    \emph{woman} is to \emph{queen}''. The model is given the first three terms as input, and then has to come up with the solution (\emph{queen}).
    Word embeddings models can, for example, apply simple linear vector arithmetic to solve the task, with $vector(man)-vector(woman)+vector(king)$.
    Then, the term closest (eg.~measured by cosine similarity) to the resulting vector is the candidate term. 
    
    In the second task type, word intrusion, the goal is to find an intruding word in a list of words, which have a given characteristic. 
    For example, find the intruder in: \emph{Austria Spain Tokyo Russia}. In our task setup, the list always includes four terms, where one
    is the intruder to be detected. 
    
    \subsection{Dataset Translation}
    \label{meth:trans}
    The given datasets are based on extended versions of English language datasets presented in~\cite{wohlgenannt2018cicling}. Those datasets were manually created inspired by categories and relations of online Wikis about ``A Song of Ice and Fire'' and ``Harry Potter''. 
    The goal was to provide high quality datasets by filtering ambiguous and very-low frequency terms. The three dimensions (2 book series, 2 task types,
    unigram and n-gram datasets) led to eight published datasets.
    
    In this work, the datasets were translated to Russian language. Two separate teams of native speakers translated the datasets to Russian. We found,
    that multiple book translations exist for the Harry Potter book series, and decided to work on two different book translations in this case. 
    As many terms, esp. named entities like book characters and location names, have a slightly different translation or transliteration from the English original to Russian, we ended up with two independent datasets. 
    
    For ASOIF, both translation teams based their translations on the same Russian book version, and there where only slight differences -- mostly regarding
    terms which are unigrams in English, but n-grams in Russian, and a few words which can have multiple translations into Russian.
    

    \subsection{Word Embedding Models}
    
       For the baseline evaluation of the datasets, we apply two popular word embedding models. Firstly, word2vec~\cite{mikolov2013w2v}
       uses a simple two-layer feed-forward network to create embeddings in an unsupervised way. The simple architecture
       facilitates training on large corpora. Basically, word similarity in vector space reflects similar contexts
       of words in the corpus. Depending on preprocessing, unigram or n-gram models can be trained. Word2vec includes
       two algorithms. CBOW predicts a given word from the window of surrounding words, while skip-gram (SG) predicts
       surrounding words from the current word. Secondly, FastText~\cite{bojanowski2016} is based on the skip-gram model,
       however, in contrast to word2vec, it makes use of sub-word information, and represents words as bag of character n-grams. 
       
       Hyperparameter tuning has a large impact on the performance of embedding models~\cite{Sahlgren2016emnlp}. 
       In the evaluations, we compare the results for different parameters settings for both datasets, 
       details on those settings are found in Section~\ref{eval:setup}.

    \subsection{Implementation}
    As mentioned, two teams worked independently on dataset translation and model training, which leads to the provision of two independent 
    GitHub repositories\footnote{\url{https://github.com/DenisRomashov/nlp2018_hp_asoif_rus}}\footnote{\url{https://github.com/ishutov/nlp2018_hp_asoif_rus}}. The repositories
    can be used to reproduce the results, and to evaluate alternative methods 
    -- based on the book translation used.
    All library requirements, usage, and evaluation, and most importantly the datasets, are found in the repositories. For model 
    creation and evaluation the popular Gensim library is used~\cite{gensim}. 
    The implementation contains two main evaluation modules, one for the \emph{analogies} task, and one for \emph{word intrusion}. In the repository,
    word intrusion is coined \emph{doesn't-match}, as this is the name of the respective Gensim function.
    Third parties can either reuse the provided evaluation scripts on a given embedding model, or use the datasets directly.
    The dataset format is the same as in word2vec~\cite{mikolov2013w2v} for \emph{analogies}, and for the word intrusion task
    it is simple the understand, with the 4 words of the task unit, and the intruder marked.

\section{Evaluation}
\label{sec:evaluation}

\subsection{Evaluation Setup}
    \label{eval:setup}
    
        \subsubsection{Book Corpora and Dataset Translation}
        
        The models analyzed in the evaluations are trained on two popular fantasy novel corpora, ``A Song of Ice and Fire'' (ASOIF) by GRR Martin, and
        ``Harry Potter'' (HP) by JK Rowling. 

        
        From ASOIF, we took the first four books; the corpus size is 11.8 MB of plain text, with a total of 10.5M tokens (11.1M before preprocessing). 
        The book series includes a large world with an immense number of characters, whereby about 30-40 main characters exists. Narration is mostly linear and
        the story is told in first person from the perspective of different main characters.
        
        The HP book series consists of seven books, with a size of 10.7 MB and 9.2M tokens (9.8M before preprocessing). 
        The books tell the story of young Harry Potter and his friends in a world full of magic. 
        The complexity of the world, and the number of characters, is generally lower than in ASOIF.
        
        The basics of dataset translations were already mentioned in Section~\ref{meth:trans}. Two independent teams worked on the task.
        In case of ASOIF, both teams based their translation work and also model creation on the same Russian book corpus.
        For HP an original translation into Russian exists, which is still the most popular one. Later other translations emerged.
        In order to cover a wider range of corpora, and to investigate the differences between those translations, the teams used 
        two different translations. The exact book versions are listed in the respective GitHub repositories\footnote{\url{github.com/ishutov/nlp2018_hp_asoif_rus/blob/master/Results.md}}\footnote{\url{github.com/DenisRomashov/nlp2018_hp_asoif_rus/blob/master/RESULTS.md}}. 
     
       \subsubsection{Preprocessing}
       
        In principle, we tried to keep corpus preprocessing to a minimum, and did only the following common steps:
        removal of punctuation symbols (except hyphens), removal of lines that contain no letters (page numbers, etc.), and sentence splitting. 
        However, in comparison to the English original version of datasets and corpus, for Russian we found that term 
        frequencies of the terms in the datasets were significantly lower. A substantial amount 
        of dataset terms even fell below the \texttt{min\_count} frequency used in model training and was thereby excluded from
        the models. This can be attributed to the rich morphology of Russian language, and other reasons elaborated 
        in the discussion section (Section~\ref{sec:dissc}).
        For this reason, we created a second version of the corpora with lemmatization applied to all tokens of the book series\footnote{Using this toolkit: \url{tech.yandex.ru/mystem}}.
        In the evaluations, we present and compare results of both corpora versions, with and without lemmatization applied.
       
        Furthermore, for the creation of n-gram annotated corpora the \emph{word2phrase} tool included in the word2vec toolkit~\cite{mikolov2013w2v} was utilized.

        \subsubsection{Models and Settings}
        
        As mentioned, we train word2vec and FastText models on the book corpora -- using
        the Gensim library.
        In the upcoming evaluations, we use the following algorithms and settings to train models: 
       




        \begin{description}
        \item[w2v-default:] This is a word2vec model trained with the Gensim default settings: 
        100-dim.~vectors, word window size and minimum number of term occurrence are both set to $5$, 
        iter (number of epochs): 5, CBOW. 
        
        \item[w2v-SG-hs]: Defaults, except: 300dim.~vectors, 15 iterations, number of negative samples: 0, 
        with hierarchical softmax, with the skip gram method\footnote{size=300, -negative=0, sg=1, hs=1, iter=15}.
        \item[w2v-SG-hs-w12:] like  \emph{w2v-SG-hs}, but with a word window of 12 words.
        \item[w2v-SG-ns-w12:] like  \emph{w2v-SG-w12}, with negative sampling set to 15 instead of hierarchical softmax.
        \item[w2v-CBOW:] like  \emph{w2v-SG-hs}, but with the CBOW method instead of skip-gram. 
        
         \item[FastText-default:] A FastText model trained with the Gensim default settings: 
        those are basically the same default settings as for \emph{w2v-default}, except for FastText-specific parameters.
        \item[FastText-SG-hs-w12:] Defaults, except: 300dim.~vectors, 15 iterations, a word-window of 12 words, number of negative samples: 0, with hierarchical softmax and the skip-gram method\footnote{size=300, -negative=0, sg=1, hs=1, iter=15, -window=12}.
        \item[FastText-SG-ns-w12:] like \emph{ft-SG-hs-w12}, but with negative sampling (15 samples) instead of hierarchical softmax. 
 \end{description}       
 
        \subsubsection{Datasets}
        
             \begin{table}
            \caption{Number of tasks and dataset sections (in parentheses) in the Russian language datasets -- with the dimensions of task
                    type, book corpus and unigram/n-gram}
            \label{tab:ds}
            \begin{center}
            \begin{tabular}[h]{cc|cc}
            \hline\noalign{\smallskip}
            
             \textbf{Book Corpus} & \textbf{Task Type} & \textbf{Unigram} & \textbf{N-Gram}   \\
            
            \noalign{\smallskip}\hline\noalign{\smallskip}
              \multirow{2}{*}{HP}       &  Analogies      &   4790 (17)  &   92 (7)  \\ 
                                       &  Word Intrusion &   8340 (19)  & 1920 (7)  \\ \hline 
            
             \multirow{2}{*}{ASOIF}    &  Analogies      &  2848  (8)   &  192 (2)  \\ 
                                       &  Word Intrusion &  11180 (13)  & 2000 (7)  \\ 
                                      
            \noalign{\smallskip}\hline
            \end{tabular}
            \end{center}
            \end{table}
      
        In total, we provide eight datasets. This number stems from three datasets dimensions: the task type (analogies and word intrusion, 
        the two book series, and the distinction between unigram and n-gram datasets). Table~\ref{tab:ds} gives an overview of the number of tasks
        within the datasets, and also of the number of sections per task. Sections reflect a subtask with specific characteristics and difficulty,
        for example analogy relations between \emph{husband} and \emph{wife}, or between between a \emph{creature} (individual) and its \emph{species}. Typically, per section,
        the items on a given side of the relation are members of the same word or named entity category, therefore a distributional language model can be more deeply analyzed for its performance in those subtasks.
        
        In contrast to popular word similarity datasets like WordSim-353~\cite{finkelstein2001placing} or MEN~\cite{Bruni2014}, most of the dataset terms are 
        named entities. Herbelot~\cite{Herbelot2015darcy} studies some of the properties of named entities in distributional
        models. For example, in the unigram word intrusion dataset, only around 7\% (ASOIF) and 17\% (HP) of terms are \emph{kinds}, the rest are named entities.
        

 \subsection{Evaluation Results}
 \label{sec:evalres}
     In this section, we present and analyze the evaluation results for the presented datasets using word embedding models.
    We start with an overview of the results of the \emph{analogies} and \emph{word intrusion} tasks, followed by more fine-grained
    results for the different subtasks of the \emph{analogies} tasks. For the \emph{word intrusion} task, we investigate evaluation
    results depending on task difficulty, and finally, a summary of results on n-gram datasets is presented.
     Further details on the results can be found on github\footnote{\url{github.com/ishutov/nlp2018_hp_asoif_rus/blob/master/Results.md}}\footnote{\url{github.com/DenisRomashov/nlp2018_hp_asoif_rus/blob/master/RESULTS.md}}. 
 
                \begin{table}[htb]
                \caption{Overall \emph{analogies} accuracy of the Russian unigram datasets for both book series. 
                Results are given for models with and without lemmatization of the corpora.
                Values for English given in parenthesis.}
                
                \label{tab:overall:an}
                \begin{center}
                \begin{tabular}{c|c|c|c|c}
                \hline\noalign{\smallskip}
            
                     Book series   & \multicolumn{2}{|c|}{ASOIF}  & \multicolumn{2}{|c}{HP} \\ 
                     Preprocessing & Minimal & Lemmatization  & Minimal & Lemmatization  \\

                \noalign{\smallskip}\hline\noalign{\smallskip}

            w2v-default                &   0.57 (8.15)  &   2.35 (-) & 00.42 (6.88) & 1.75  (-)   \\
            w2v-SG-hs                  &  17.61 (28.44) &  24.40 (-) & 13.40 (25.11) & 23.00 (-)  \\
            w2v-SG-hs-w12              &  \textbf{24.56} (37.11) &  \textbf{32.66} (-) & \textbf{20.34} (30.00) & \textbf{28.95} (-)  \\
            w2v-SG-ns-w12              &  21.58 (29.32) &  20.97 (-) & 13.17 (20.84) & 12.99 (-)  \\
            w2v-w12-CBOW               &   0.57 (2.67)  &   1.07 (-) &  0.68 (7.22) &  2.62 (-)  \\

            FastText-default           &   0.42 (1.33) &    2.31 (-) &  0.08 (0.87) &  0.42 (-)  \\
            FastText-SG-hs-w12         &  11.04 (29.81) &  21.58 (-) &  8.57 (25.46) & 19.30 (-)  \\
            FastText-SG-ns-w12         &   0.99 (14.64) &  0.8   (-) &  3.77 (14.23) & 4.13  (-)  \\

                \noalign{\smallskip}\hline
                \end{tabular}
                \end{center}
            \end{table}

    Table~\ref{tab:overall:an} provides an overview of results for the \emph{analogies} task. It includes the results for the two book series ASOIF
    and HP. We distinguish two types of input corpora, namely with and without the application of lemmatization (``minimal'' preprocessing vs.~``lemmatization'').
    Embedding models were trained on the corpora with the settings described in Section~\ref{eval:setup}.
    Furthermore, the evaluation scores for English language corpora and datasets are given for comparison.

        The results in Table~\ref{tab:overall:an} indicate that models trained with the skip-gram algorithm clearly outperform CBOW 
        for analogy relations. Another important fact is that for Russian language lemmatization of the corpus tokens before training
        has a strong and consistent positive impact on results. However, the numbers for Russian stay below the numbers for English.
        Both the performance impact of lemmatization, and the differences between English may partly be the result of differences in
        corpus term frequency. This intuition will be investigated and discussed in Section~\ref{sec:dissc}.


                \begin{table}[htb]
                \caption{Overall \emph{word intrusion} accuracy (in percent) for the Russian unigram datasets for both book series. 
                Results are given for models with and without lemmatization of the corpora.
                Values for English given in parenthesis.}
                \label{tab:overall:wi}
                \begin{center}
                \begin{tabular}{c|c|c|c|c}
                \hline\noalign{\smallskip}
            
                     Book series   & \multicolumn{2}{|c|}{ASOIF}  & \multicolumn{2}{|c}{HP} \\ 
                     Preprocessing & Minimal & Lemmatization  & Minimal & Lemmatization  \\

                \noalign{\smallskip}\hline\noalign{\smallskip}

            w2v-default                &  62.03 (86.53) &  64.83 (-)  & 34.69 (64.83) & 53.59 (-)   \\
            w2v-SG-hs                  &  65.93 (77.9)  &  \textbf{73.30} (-)  & 55.99 (73.3)  & 60.44 (-)  \\
            w2v-SG-hs-w12              &  67.11 (74.86) &  68.89 (-)  & \textbf{61.09} (68.69) & 59.87 (-)  \\
            w2v-SG-ns-w12              &  \textbf{68.09} (75.15) &  67.1  (-)  & 58.43 (74.43) & 57.01 (-)  \\
            w2v-w12-CBOW               &  57.35 (75.61) &  61.28 (-)  & 42.39 (61.28) & 48.73 (-)  \\

            FastText-default           &  61.59 (73.82) &  56.56 (-)  & 41.92 (56.56) & 46.50 (-)  \\
            FastText-SG-hs-w12         &  66.82 (75.99) &  70.20 (-)  & 60.99 (70.2)  & 60.27 (-)  \\
            FastText-SG-ns-w12         &  67.81 (75.38) &  68.41 (-)  & 59.13 (76.41) & \textbf{61.54} (-)  \\ \hline

            Stock Embeddings           &   27.3 (-) &  - (-) &  25.36 (-) &  - (-)  \\
            Random Baseline            &   25.00  &  25.00  &  25.00 &  25.00   \\
                \noalign{\smallskip}\hline
                \end{tabular}
                \end{center}
            \end{table}
            
    Table~\ref{tab:overall:wi} gives the results for the \emph{word intrusion} tasks. Again, we distinguish between preprocessing with and
    without lemmatization, and between the results of different models trained on the two book series. For the word intrusion task, the differences 
    observed between skip-gram and CBOW, and regarding lemmatization, are smaller as compared to 
    \emph{analogies} results in Table~\ref{tab:overall:an}; however, the tendencies still exist.
    For comparison, we also applied pretrained FastText models (``Stock Embeddings'') trained on Wikipedia\footnote{\url{https://github.com/facebookresearch/fastText/blob/master/pretrained-vectors.md}} to the task.
    As expected, those models perform very poorly, only slightly over the random baseline.

  
            \begin{table}[htb]
                \caption{ASOIF Analogies Russian dataset: Accuracy of different word embedding models on selected analogies task sections, and total accuracy.
                In parenthesis, values from the English language dataset are given for comparison.}
                \label{tab:sections}
                \begin{center}
                \begin{tabular}{c|c|c|c|c|c}
                \hline\noalign{\smallskip}
                      Task Section & first-lastname & husband-wife  & loc-type & houses-seats & Total \\
                        \noalign{\smallskip}\hline\noalign{\smallskip}
            Number of tasks:       & 2368 & 30 & 168 & 30 &  2848 \\ 
            
                        \noalign{\smallskip}\hline\noalign{\smallskip}
            w2v-default                &  1.93  (8.78) &   0.0  (6.67) & 5.0  (4.76)  & 10.0  (20.0)  & 2.35  (8.15)    \\
            w2v-SG-hs                  &  26.98 (32.01) &   15.0 (10.0) & 7.5  (11.9)  & 26.67 (40.0)  & 24.4  (28.44)    \\
            w2v-SG-hs-w12              &  \textbf{36.24} (42.36) & \textbf{20.0} (10.0) & 6.25 (19.64)  & \textbf{30.0}  (33.33)  & \textbf{32.66} (37.71)    \\
            w2v-SG-ns-w12              &  32.62 (40.62) &   5.0  (6.67) & \textbf{12.5} (22.62)  & 26.67 (40.0)  & 29.62 (36.41)    \\
            w2v-w12-CBOW               &  0.69  (1.27) &   5.0  (6.67) & 2.5  (11.9)  & 6.67  (30.0)  & 1.07  (2.67)    \\
            FastText-default           &  2.33  (1.06) &   5.0  (3.33) & 0.0  (3.57)  & 3.33  (3.33)  & 2.31  (1.33)    \\
            FastText-SG-hs-w12         &  23.86 (34.04) &   15.0 (6.67) & 6.25 (13.69)  & 20.0  (46.67)  & 21.58 (29.81)    \\
            FastText-SG-ns-w12         &  27.33 (35.09) &   15.0 (3.33) & 5.0  (11.9)  & 13.33 (26.67)  & 24.4  (30.44)    \\

                \noalign{\smallskip}\hline
                \end{tabular}
                \end{center}
            \end{table}
            
        All datasets are split into various sections, which reflect specific relation types, for example~\emph{child-father} or \emph{houses-and-their-seats}.
        Those relations have certain characteristics, such as involving person names, location entities, or other, which allow a fine-grained analysis and comparison of embedding models and their performance. Table~\ref{tab:sections} shows some selected sections from the ASOIF analogies dataset.
        The performance varies strongly over the different subtasks, but the data indicates, that models that do well in total, are also more suitable on the individual tasks.

            \begin{table}[ht]
                \caption{Accuracy results with regards to task difficulty -- Russian ASOIF word intrusion dataset (unigrams), trained on a lemmatized corpus.} 
                \label{tab:diff:over:td}
                \begin{center}
                \begin{tabular}{c|c|c|c|c|c}
                \hline\noalign{\smallskip}
                      Task Difficulty & 1 (hard)  & 2 (med-hard)  & 3 (medium) & 4 (easy) & AVG \\
                        \noalign{\smallskip}\hline\noalign{\smallskip}
                      Number of tasks:       & 2795 & 2795 &  2795 & 2795 & 11180 \\
                        \noalign{\smallskip}\hline\noalign{\smallskip}
                w2v-default          &  61.82 & \textbf{72.9}  & 70.16 & 91.74 & \textbf{74.17} \\ 
                w2v-SG-hs            &  46.27 & 67.72 & 73.38 & 85.33 & 68.18 \\ 
                w2v-SG-hs-w12        &  39.18 & 67.73 & \textbf{75.67} & 85.83 & 67.1  \\
                w2v-SG-ns-w12        &  57.53 & 70.98 & 74.35 & 90.84 & 73.43 \\
                FastText-default     &  \textbf{61.86} & 65.62 & 66.26 & \textbf{96.85} & 72.65 \\
                FastText-SG-hs-w12   &  46.76 & 67.59 & 73.38 & 87.48 & 68.8  \\
                FastText-SG-ns-w12   &  39.28 & 66.87 & 73.56 & 86.37 & 66.52 \\

                \noalign{\smallskip}\hline
                \end{tabular}
                \end{center}
                \end{table}
                
            The word intrusion datasets were created with the idea of four task difficulty levels. The \emph{hard} level includes near misses; on the \emph{medium-hard} level, the outlier still has some semantic relation to the target terms, and is of the same word (or NE) category.
            On the \emph{medium} level outliers are of the same word category, but have little semantic relatedness to the target term. And finally, in the \emph{easy} category, the terms have no specific relation to the target terms. 
            As an example, if the target terms are~\texttt{Karstark Greyjoy Lannister}, ie.~names of \emph{houses}, then a \emph{hard} intruder
            might be \texttt{Theon}, who is a person from one of the houses. \texttt{Bronn} will be a \emph{med-hard} intruder, also a person, not from those houses.
            \texttt{Winterfell} (a location name) will be in the \emph{medium} category, and \texttt{raven} in the easy one.
            
            Very interestingly, models using the CBOW algorithm (such as \emph{w2v-default} and \emph{FastText-default}) provide very good results on the hardest task category with over 60\% of correct intruders selected. 
            On the other hand, SG-based models only show 39\%-46\%. 
            For the easiest category, \emph{FastText-default} excels with almost 97\% accuracy.
            In general word embeddings, esp.~when trained on small datasets and using cosine similarity for the task, struggle to single out terms in the \emph{hard} category by a specific (minor) characteristic of the target terms. This will be further discussed in section~\ref{sec:dissc}.

        \subsubsection{N-Gram Results}
        As mentioned, in addition to the four unigram dataset, we created complementary n-gram datasets for the two book series and the two task types.
        In correspondence with the n-gram detection method used~\cite{mikolov2013w2v}, in the datasets n-grams are words connected by the underscore symbol.
        Many of the terms are person or location names such as \emph{Forbidden\_Forest} or \emph{Maester\_Aemon}. For reasons of brevity, we will not 
        include result tables here (see GitHub for details). The general tendency is that n-gram results are below unigram results in the \emph{analogies} task,
        for word intrusion results are comparable with around 70\% accuracy (depending on model settings).
        In comparison with word2vec, FastText-based models perform better on n-gram than unigram tasks, 
        this can be explained by the capability of FastText to leverage subword-information within n-grams. 

\section{Discussion}
\label{sec:dissc}

In general, there is quite a big difference in performance between Russian and English datasets and models, when the same (minimal) preprocessing is being applied to the corpora.
For example, in Table~\ref{tab:overall:an} the best ASOIF performance for Russian (with minimal preprocessing) is 24.56\%, but 37.11\% for English, and for HP the values are
20.34\% for Russian, and 30.00\% for English. In the case of \emph{word intrusion}, the same pattern repeats: 68.09\% for Russian ASOIF, and 86.53\% for English, 
and finally, 61.09\% for the Russian HP dataset versus 74.43\% (English).

Our first intuition was, that the rich morphology of the Russian language, where also proper nouns have grammatical inflections by case, might reduce the
frequency of dataset words in the corpora. Sahlgren and Lenci~\cite{Sahlgren2016emnlp} show the impact of term frequency on task accuracy in the general domain.
Subsequent analysis shows, that eg.~for the HP dataset, the average term frequency of dataset terms is 410 for the English terms and English book corpus, while for
the Russian it is only 249. We then decided to apply lemmatization to the Russian corpora, which helped to raise Russian average term frequency to 397.
Also the evaluation results improved overall, as seen in the tables in Section~\ref{sec:evalres}. However, despite the positive effects, lemmatization
also introduces a source of errors. For some dataset terms, the frequency even becomes lower; after lemmatization, the number of Russian dataset
terms that are below the \emph{min\_count} threshold to be introduced into the word embedding models rises. An example of such problem cases is
the word ``Fluffy'' from HP, which was translated as ``Пушок''. But then the lemmatizer wrongly changed it to ``пушка'' (a gun), so that ``Пушок'' disappeared from the trained models.

A number of other difficulties and reasons for the lower performance on the Russian datasets emerged: a) 
When comparing the output of the two translation teams, we observed words that have many meaningful translations to Russian, thereby lowering term frequency. For example, ``intelligence'' can be translated as ``ум'', ``интеллект'', ``осознание'', ``остроумие'' and so on. 
b) The transliteration of English words into Russian is not always clear,
and we have found in the analysis that even within the same book corpora (translations) it is not always consistent, 
even more so between different translators.
For example, there is no [æ] phonetic sound in Russian, so it can be transliterated to multiple letters: \emph{а}, \emph{е}, \emph{э}.
c) In Russian, the \emph{ё} letter is often replaced with \emph{е}. If this happens inconsistently, it impacts term frequencies.


Another interesting aspect is the performance of the models on various difficul\-ty levels in the word intrusion tasks.
If difficulty is low, then common word embedding models already work very well, in our experiments with an accuracy up to 97\%.
However, in the \emph{hard} category terms are very similar in their overall semantics and context, but the target terms possess one characteristic that the 
intruder lacks. Cosine similarity just looks at overall vicinity in vector space, with a success rate of ca.~40--60\%.
There has been some work on named entities and the distinction between \emph{individuals} (and their properties) and \emph{kinds} within distributional models~\cite{Herbelot2015darcy,Boleda2017iwcs,Boleda2017eacl}, but there is still much room on how to tackle such issues in a general way. 


\section{Conclusions}
\label{sec:concl}

In this work, we present Russian language datasets in the digital humanities domain for the evaluation of distributional semantics models. 
The datasets cover two basic task types, \emph{analogy} relations and \emph{word intrusion} for two well-known fantasy novel book series.
The provided baseline evaluations with word2vec and FastText models show that models for the Russian versions of the corpora and datasets offer lower accuracy than for the English originals. 
The contributions of the work include: a) the translation to Russian and provision (on GitHub) of eight datasets in the digital humanities domain, b) providing baseline evaluations and comparisons for various settings of popular word embedding models, c) studying the effects of preprocessing (esp.~lemmatization) on performance, d) analyzing the reasons for differences between Russian
and English language evaluations, most notably term frequency and issues arising from translation.

\section{Acknowledgments}
This work was supported by the Government of the Russian Federation (Grant 074-U01) through the ITMO Fellowship and Professorship Program. 

\bibliographystyle{splncs}
{\small \bibliography{myjabref}}

\end{document}